\documentclass[10pt,twocolumn,letterpaper]{article}

\usepackage{cvpr}              

\usepackage{graphicx}
\usepackage{amsmath}
\usepackage{amssymb}
\usepackage{booktabs}
\usepackage{cite}
\usepackage{makecell}
\usepackage{array}
\usepackage{url}
\usepackage{xcolor}
\usepackage{caption}
\usepackage{changepage}
\usepackage[pagebackref,breaklinks,colorlinks]{hyperref}
\usepackage[capitalize]{cleveref}

\crefname{section}{Sec.}{Secs.}
\Crefname{section}{Section}{Sections}
\Crefname{table}{Table}{Tables}
\crefname{table}{Tab.}{Tabs.}

\newcommand{\tablestyle}[2]{\setlength{\tabcolsep}{#1}\renewcommand{\arraystretch}{#2}\centering\footnotesize}

\begin{document}

\title{Improvements to Self-Supervised Representation Learning for Masked Image Modeling}

\author{Jiawei Mao$^{1}$   \quad Xuesong Yin$^{1*}$ \quad Yuanqi Chang$^{1}$ \quad Honggu Zhou$^{1} \quad $  \\ 
$^{1}$School of Media and Design, Hangzhou Dianzi University, China \qquad \\
{\tt\small\{jiaweima0,yinxs,211330020,211330022\}@hdu.edu.cn }\\
}

\maketitle

\begin{abstract}
This paper explores improvements to the masked image modeling (MIM) paradigm. The MIM paradigm enables the model to learn the main object features of the image by masking the input image and predicting the masked part by the unmasked part. 
We found the following three main directions for MIM to be improved. First, since both encoders and decoders contribute to representation learning, MIM uses only encoders for downstream tasks, which ignores the impact of decoders on representation learning. 
Although the MIM paradigm already employs small decoders with asymmetric structures, we believe that continued reduction of decoder parameters is beneficial to improve the representational learning capability of the encoder. 
Second, MIM solves the image prediction task by training the encoder and decoder together, and does not design a separate task for the encoder. 
To further enhance the performance of the encoder when performing downstream tasks, we designed the encoder for the tasks of comparative learning and token position prediction. 
Third, since the input image may contain background and other objects, and the proportion of each object in the image varies, reconstructing the tokens related to the background or to other objects is not meaningful for MIM to understand the main object representations. 
Therefore we use ContrastiveCrop to crop the input image so that the input image contains as much as possible only the main objects. 
Based on the above three improvements to MIM, we propose a new model, Contrastive Masked AutoEncoders (CMAE). We achieved a Top-1 accuracy of 65.84\% on tiny-imagenet using the ViT-B backbone, which is +2.89 outperforming the MAE of competing methods when all conditions are equal. 
Code will be made available.
\end{abstract}

\section{Introduction}
\label{sec:intro}

Over the years vision transformers have shown the ability to rival deep convolution in the field of computer vision by focusing on the similarity between its own elements within an image to improve the semantic information of the representation. 
However, the training of the transformer in the image domain is more difficult. First, There are natural differences between images and natural language e.g., images are continuous, natural language is discrete. 
Second, The transformer is computationally expensive because it requires a global computation of the image.

In the field of self-supervised learning, the emergence of masked image modeling (MIM) has alleviated the above problems to some extent. MIM uses a auto-encoder (AE)\cite{hinton2006reducing} architecture to train the transformer. 
It greatly reduces the computation of transformer by masking the input image token. MIM uses unmasked tokens to predict masked tokens, allowing the transformer model to grasp richer semantic information of the image at the encoding stage.

We propose that Contrastive Masked AutoEncoders (CMAE) aims to design a better paradigm to improve transformer performance based on MIM. 
We have studied MIM and its variants in detail, and pointed out the following three directions to improve MIM:

(i) We find that the MIM model achieves representation learning by the encoder and decoder working together for mask token prediction. 
However, the purpose of MIM is to make the encoder learn a good representation for the next downstream tasks during the prediction process. However, the presence of the decoder inevitably weakens the performance of the encoder. 
Although MIM has been initially designed with an asymmetric structure of weak decoder for the mask token reconstruction task.  But we found in \cite{xie2021simmim} that the decoder can be reduced to a very low number of parameters. 
We therefore believe that the capability of the encoder can be indirectly improved by continuing to weaken the decoder while maintaining the reconstruction token.

(ii) The loss function of MIM is a common constraint on the encoder and decoder and does not provide an independent loss function for the encoder. 
Therefore the feature extraction capability of the encoder can be enhanced by adding new tasks to it. In CMAE, we designed two new tasks independently for the encoder to improve its performance during downstream tasks. 
\cite{chen2022context} separately visualizes the regions that the MIM and the contrast learning model focus on by performing two different tasks separately. We found that the encoder in MIM focuses more on the image global by reconstructing the image. 
Contrastive learning obtains unique features of an image by bringing positive samples closer and pushing negative samples farther in a high-dimensional space. Therefore, we designed the two tasks as one. 
CMAE expects the encoder to perform image prediction tasks in a way that the features it extracts contain both global and image-specific information by means of contrast learning. 
In addition, we found a focus on linguistic order in many natural language models that improve on BERT \cite{devlin2018bert}, such as ALBERT \cite{lan2019albert}. While MIM uses random masking, it gets the order of unmasked tokens in a random order. 
In order to enable the encoder to independently grasp the spatial location information of each token, we designed a location prediction task for it. 
This task can also be understood as a disorder recovery task since the correct position of the token is mastered to convert the disorder to order.

(iii) We consider that the input image may also have an impact on the MIM. Generally the input image contains not only the main object but also the background and other objects, and all objects in the image are rendered at different scales. 
When the proportion of primary objects is relatively small or the unmasked tokens are mainly primary objects, the reconstruction of background or other objects is not meaningful for the encoder to learn the primary object features. 
Also, to perform the contrastive learning task better, we introduce ContrastiveCrop \cite{peng2022crafting} to process the input images. 
It helps to crop the primary object area by positioning the primary object with the semantic information extracted by the encoder, thus minimizing the interference of background and secondary objects.

The main contributions of this paper can be summarized as:

\begin{itemize}
  \item We discuss the use of a simpler architecture for the MIM decoder to indirectly improve the feature extraction capability of the encoder in MIM. We found that properly weakening the decoder helps the MIM perform downstream tasks while maintaining the recovered image quality.
\item CMAE makes the encoder perform the contrastive learning task while the MIM model performs the image recovery task. This enables the encoder to extract features that focus on both global and image-specific attributes.
\item Inspired by the improvement of Masked Language Model (MLM) in NLP, we designed the position prediction task for the encoder. This task helps the encoder to have information about the spatial location of each token in the image.
\item We consider the effect of the input image on the MIM. By introducing ContrastiveCrop into MIM, it makes MIM more focused on the main object and provides help for contrastive learning in CMAE.
\end{itemize}

\begin{figure*}[h]
  \centering
   \includegraphics[width=1\linewidth]{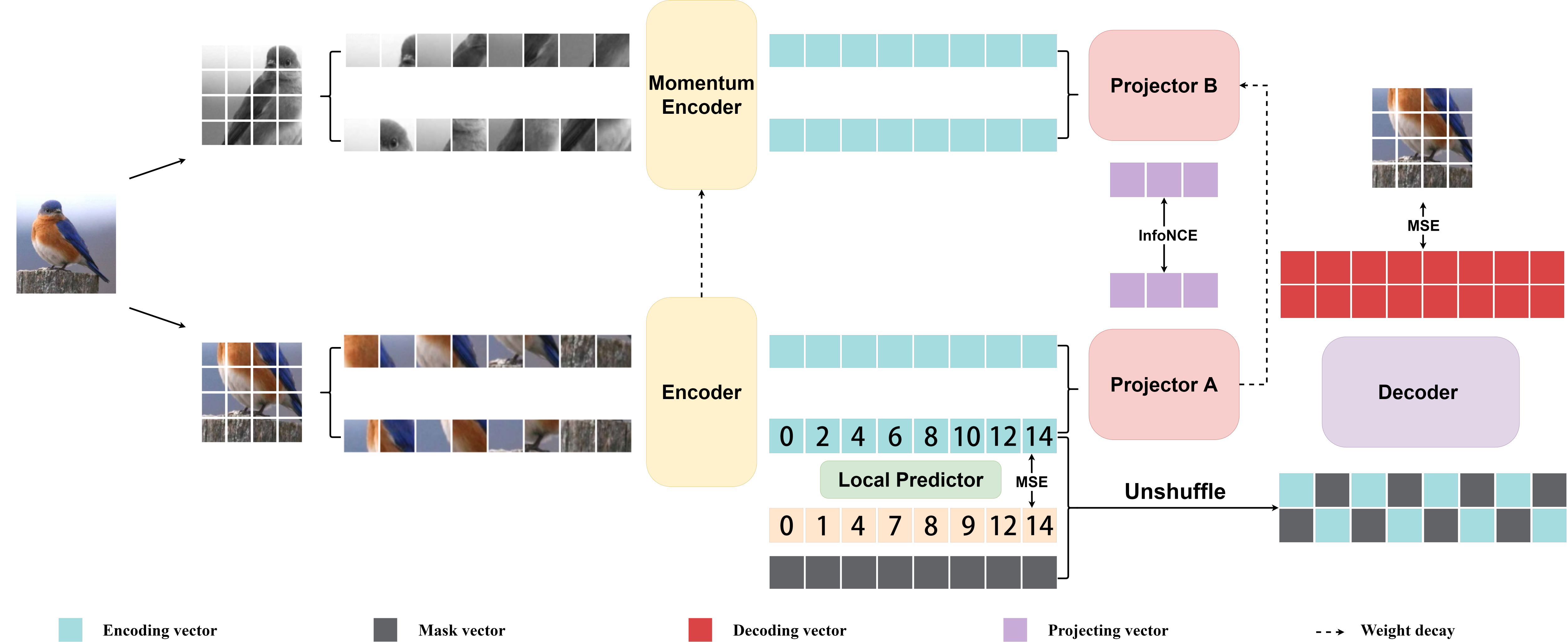}
   \caption{Our CMAE architecture. Our CMAE introduces ContrastiveCrop for MIM to optimize the input data of MIM. Meanwhile, we design independent tasks such as comparison learning and position prediction for the encoder of MIM.}
   \label{fig1}
\end{figure*}

\section{Related Work}

\textbf{Contrastive self-supervised learning.} The notion of contrastive learning has been proposed for a long time, but it was Chen et al.'s SIMCLR \cite{chen2020simple} that really made it popular. 
The framework extracts representations that substantially outperform current SOTA, sparking a boom in contrastive self-supervised learning. 
He et al. proposed MOCO v1 \cite{he2020momentum} and introduced the idea of queue in/out and encoder momentum update into the comparative learning, which solved the limitation of GPU memory on batch size. 
With the introduction of SIMCLR, He et al. improved the data augment method based on MOCO v1 by adding a projection layer after the encoder and using cosine decay to create a more powerful MOCO v2 \cite{chen2020improved}. 
Chen et al. also improved SIMCLR by adopting a deeper ResNet-152 as the encoder and using a deeper projection layer, updating it to SIMCLR v2 \cite{chen2020big}. 
Since comparative learning requires many negative examples for comparison, which is both time and memory consuming, Caron et al. proposed SwAV \cite{caron2020unsupervised} to reduce the loss of comparative learning by means of clustering. 
Caron et al. then proposed comparative learning training on larger open domain data, leading to the SEER \cite{goyal2021self}. 
Grill et al. proposed BYOL \cite{grill2020bootstrap}, which solves the pattern collapse problem associated with contrastive learning by stopping the gradient and adding batch normalization to the projection layer in a way that allows the model to be compared implicitly. 
By exploring the Siamese Network, He et al. found that stopping the gradient is is the key to avoid pattern collapse, and proposed SimSiam \cite{chen2021exploring}. To better train the encoders in the MIM model, we introduce a pre-training task of contrastive learning for CMAE.

\textbf{Generative self-supervised learning.} Generative self-supervised models are mainly auto-encoder and its variants, Hinton et al. were the first to propose a auto-encoder (AE) \cite{hinton2006reducing} for the purpose of dimensionality reduction of data. 
To reduce computational consumption, Bengio et al. proposed a sparse auto-encoder \cite{vincent2010stacked}. Kingma et al. designed the powerful generative model of variational auto-encoder (VAE) \cite{kingma2013auto} by extrapolating the variational Bayes. 
Vincent et al. proposed Denoising autoencoder (DAE) \cite{gidaris2019generating} to mitigate the overfitting problem of AE by adding noise to the input, which improved the robustness of AE. 
With the popularity of convolutional neural networks (CNN) \cite{krizhevsky2012imagenet,he2016deep}, Masci et al. proposed a convolutional auto-encoder (CAE) \cite{chen2022context} based on the AE, Oord et al. created the Vector QuantisedVariational AutoEncoder (VQ-VAE) \cite{razavi2019generating}, which produces very realistic images, by making the encoder output discrete. 
Recently, He et al. proposed a new auto-encoder called Masked AutoEncoders \cite{he2021masked} for training transformer by partially masking the image. 
chen et al. designed the context autoencoder \cite{chen2022context} by separating the two tasks of image reconstruction and representation learning. Dong et al. created Peco \cite{dong2021peco} by combining MIM with a discrete codebook. 
Xie et al. have also achieved good results in image recovery and downstream tasks using fully connected layers as decoders in simMIM \cite{xie2021simmim}. These related works have helped us tremendously in studying the improvements to MIM.

\textbf{Transformer.} The attention mechanism was first proposed by Vaswani, Ashish, et al. \cite{vaswani2017attention}. It is the core of the transformer. 
In recent years, Dosovitskiy et al. proposed an efficient visual transformer by chunking images: the Vision Transformer (VIT) \cite{dosovitskiy2020image}, which achieved SOTA results in many tasks. 
To moderate the reliance of VIT on large datasets, Touvron et al. used CNNs as teachers and transformers as students, and designed Deit \cite{touvron2021training} through this use of a teacher-student strategy in pretraining. 
By knowledge distillation training strategy \cite{hinton2015distilling}, Touvron et al. also proposed CAIT \cite{touvron2021going}, which can train transformer without external data. Wang et al. proposed PVT \cite{wang2021pyramid}, which uses a pyramid structure to adapt the transformer to intensive prediction tasks. 
In order to solve the problems of varying instance sizes and large pixel computations that exist in the field of transformer in images, Liu et al. proposed Swin Transformer \cite{liu2021swin} by moving the window to compute attention. 
Our CMAE is dedicated to researching better training methods for vision transformers.

\section{Methodology}

\subsection{Contrastive Masked AutoEncoders Framework}

Our CMAE works on the improvement of MIM to exploit the greater potential of transformer in computer vision self-supervised learning (SSL). Our general framework is shown in Fig.\ref{fig1}. CMAE contains four major improvements:

\begin{itemize}
  \vspace{-5pt}
\item[1)] Data Augmentation. Given an input image, CMAE will use ContrastiveCrop for data augmentation after a certain number of epoch training. ContrastiveCrop will improve the input data of MIM.
\vspace{-5pt}
\item[2)] Contrastive Tasks. We introduce a contrastive task for the CMAE encoder to enhance the feature extraction capability of the encoder. 
Contrastive learning allows the CMAE encoder to extract features unique to the image, further enhancing the performance of MIM when performing downstream tasks.
\vspace{-5pt}
\item[3)] Location Prediction Tasks. To enable the encoder of MIM to have the spatial position information of the image, we designed the position prediction task for the encoder of CMAE.
\vspace{-5pt}
\item[4)] Decoder. Through the study of \cite{xie2021simmim} we found that the decoder in MIM can use a simpler structure.Therefore, we have experimented with various decoder architectures to find the decoder framework that can improve the performance of MIM.
\end{itemize}

\subsection{Data Augmentation}
\begin{figure}[t]
  \centering
   \includegraphics[width=1\linewidth]{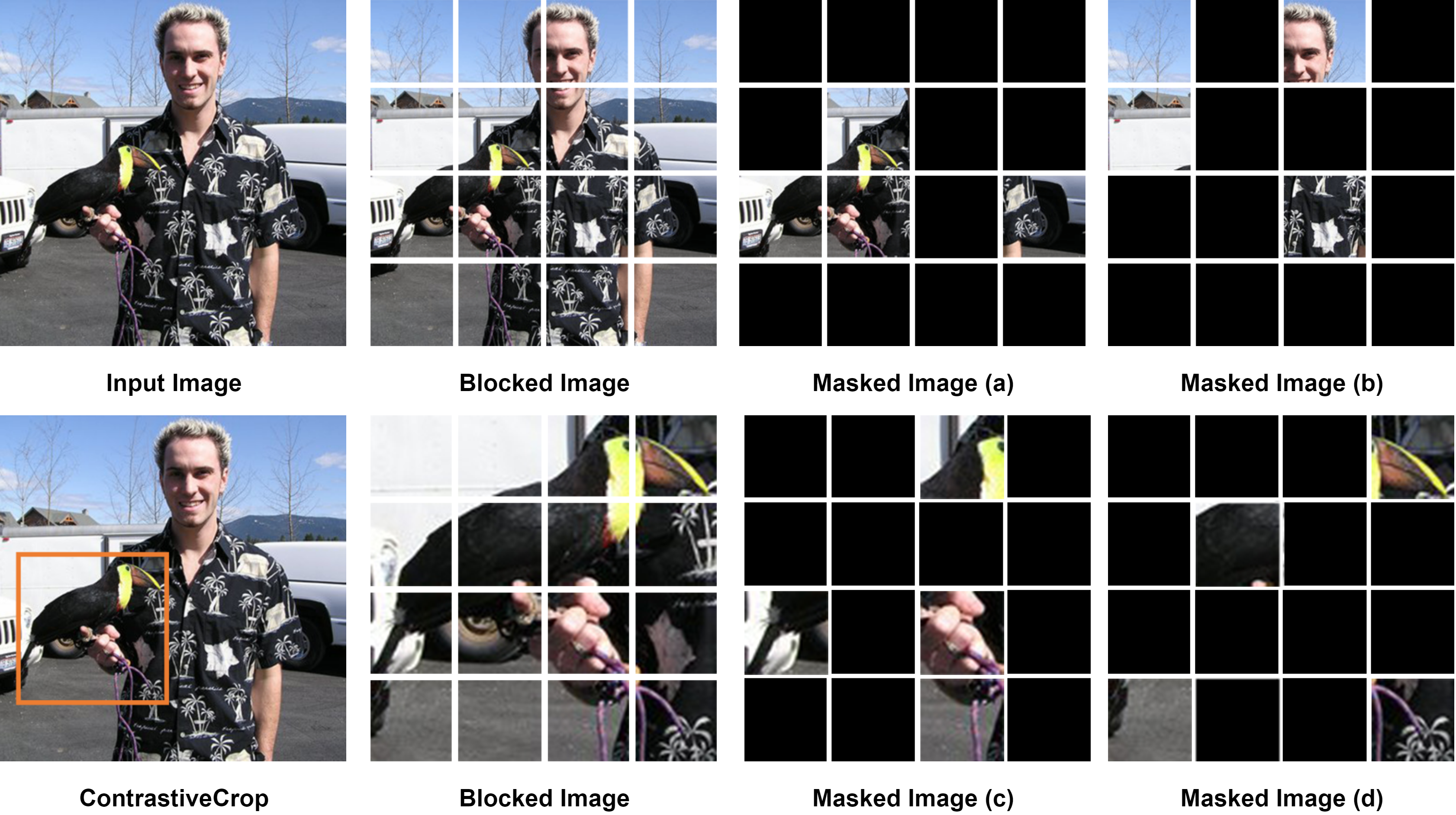}
   \caption{The first row shows two types of problems that may result from random masking of MIM input data with unmasked tokens all primary objects (a) and masked tokens all primary objects (b). The second row shows the effect of introducing ContrastiveCrop (c)(d).}
   \label{fig2}
\end{figure}

We first discuss improvement options for the MIM input. We found that when the proportion of the main object in the MIM input image is relatively small or there are more other objects may affect the performance of MIM. 
As shown in Fig.\ref{fig2}, since the MIM performs random masking, all the markers associated with the primary object may be unmasked markers when the above two cases occur Fig.\ref{fig2}(a). 
At this point the MIM performs the use of primary object tokens to predict background or other irrelevant objects. This does not achieve the purpose of the MIM to learn the primary object representations by mutual prediction between the primary object tokens. 
Or random masks may also cause the images in the above two cases to appear as in Fig.\ref{fig2}(b). It is also not logical to recover the token of the main object that is masked by the token of the unmasked background and other objects at this point. 
Therefore, we introduce ContrastiveCrop \cite{peng2022crafting} to perform data augmentation on the input images in MIM. 
Since contrastive learning draws the positive case samples closer and pushes the negative case samples farther in the high-dimensional space. This forces the model to focus on the main object to discover unique features of the image. 
As a result, ContrastiveCrop automatically locates the main objects for cropping based on the main semantic information extracted by the contrast learning model. 
Fig.\ref{fig2}(c)(d) shows the possible results of using ContrastiveCrop for data augmentation of the input image with random masking. 
We can observe that this method maximizes the proportion of the main object in the input image and greatly suppresses the background and other object information. 
When the image is basically the main object, the above two types of problems can be avoided as much as possible, thus achieving the original purpose of MIM. 
And ContrastiveCrop plays a mutually reinforcing role with the contrast learning task we added to MIM.  ContrastiveCrop usually uses the weighted result of each channel of the last layer of the feature map of the convolution layer to produce a heat map (M) to locate the main object position for cropping, which can be written as,

\begin{equation}
  \begin{split}
  &({x_1},{x_2},{x_3},{x_4}) = {R_{crop}}(s,r,I)\\
  &I = F(1[M > k])
\end{split}
  \label{eq1}
\end{equation}

where $R_{crop}$() is a random sampling function that is used to return quaternions ($x_1$, $x_2$, $x_3$, $x_4$) representing the crop, s denotes scaling, r denotes aspect ratio. k is a threshold value between 0 and 1, s, 1 is the indicator function and F is used to calculate the rectangular region I after feature map activation. 
Since we explored how to better train the transformer using MIM, while ContrastiveCrop only investigated localization using convolutional feature maps. 
Therefore, we tested the results of using the feature output of the last layer of VIT \cite{dosovitskiy2020image} and the multi-headed attention map using the last stage separately. 
The reason we consider multi-headed attention maps is that we think that attention maps in the transformer may be more suitable for heat map production of the main objects than the last layer of output. 
Because the attention map represents the most relevant regions in the feature map.

\subsection{Contrastive Tasks}

Self-supervised learning includes contrastive self-supervised learning and generative self-supervised learning. Contrastive self-supervised learning uncovers unique features of images by comparing positive and negative examples in a high-dimensional space. 
Generative self-supervised learning learns image information by reconstructing the image. MIM belongs to generative self-supervised learning in self-supervised learning. \cite{chen2022context} shows the regions of interest for MIM and contrastive learning, respectively. 
We find that MIM prefers to observe the image globally in order to meet the need for image recovery. MOCO-v3 \cite{chen2021empirical}, on the other hand, is more oriented to observe unique regions of the image in order to bring the positive examples closer and move away from the negative ones. 
Thus, in order to make MIM better train the transformer, we introduce contrastive learning as an independent task of the MIM encoder. Thereby, the transformer in CMAE can both mine the unique features of the image and retain the global information of the image. 
We first initialize two encoders of the same weight, E and ME, where ME is updated by the E momentum. 
Then two data-enhanced images $x_q$, $x_k$ belonging to the same object will be subjected to random masking operations to obtain the unmasked vector $q_1$,$k_1$ and the masked vector $q_2$,$k_2$ respectively. 
Then E will extract the features of $q_1$ and $q_2$, and ME will extract the features of $k_1$, $k_2$, but since $q_2$ belongs to the unmasked vector, CMAE will not record the gradient of $q_2$. 
Eventually, as shown in Eq.\ref{eq2}, the two types of token representations belonging to the same image are merged and the correct order is restored.

\begin{equation}
  \begin{split}
  &{{\rm{h}}_1} = restore(concat[{q_1},detach({q_2})])\\
  &{{\rm{h}}_{\rm{2}}} = restore(concat[{k_1},{k_2}])
\end{split}
  \label{eq2}
\end{equation}

where detach() means stopping the gradient operation, concat() means merging the vectors, and restore() means restoring the order. 
Finally $h_1$ and $h_2$ will be input into the projection layer P and the projection layer MP updated by the momentum of P (MP is initialized by the weights of P), respectively, to obtain the higher order vectors $z_q$, $z_k$. 
The dot product is used to measure the similarity between $z_q$,$z_k$ and InfoNCE \cite{van2018representation} is used as the loss function for the comparative task:
\begin{equation}
\begin{split}
{L_{{\rm{ctr}}}} =  - \log \frac{{\exp ({z_q}.{z_k}/\tau )}}{{\exp ({z_q}.{z_k}/\tau ) + \sum\limits_{{z_{{k^ - }}}} {\exp ({z_q}.{z_{{k^ - }}}/\tau )} }}
\end{split}
\label{eq3}
\end{equation}

Where $\tau$ is a temperature hyper-parameter \cite{wu2018unsupervised}, The set ${z_{k-}}$ consists of other image outputs of ME and MP.

\subsection{Location Prediction Tasks}

By studying algorithms such as albert \cite{lan2019albert} in NLP, we found that having spatial location information helps to improve encoder performance. 
Through further research in the field of computer vision about \cite{chen2019destruction} we found that the spatial location information is beneficial to improve the local modeling capability of the model. 
So we designed another location prediction task for the MIM encoder to enhance the transformer's understanding of the image local information. Since MIM uses a random mask, this guarantees us the feasibility of this task. Meanwhile, predicting the position of each token will restore each token to its correct position. 
Consequently, this task can also be considered as a disorder recovery task. Because we do not want the encoder to observe the masked token, CMAE only predicts the correct location of the unmasked token. As it is a random mask, each location is considered. 
As illustrated in Eq.\ref{eq4}, we use only two MLP layers for location prediction.
\begin{equation}
    {\rm{p}} = MLP(MLP({q_1}))
  \label{eq4}
  \end{equation}

p denotes the spatial coordinates of the unmasked token q1 predicted by CMAE. Note that q1 does not contain the cls token at this point because the cls token does not have a real position in the image. 
Then CMAE converts the real coordinates to the unique thermal encoding form t. The location prediction losses are as follows,

\begin{equation}
{L_{{\rm{loc}}}} = ||q - t|{|_2}
\label{eq5}
\end{equation}

\subsection{Decoder Framework}

Motivated by \cite{xie2021simmim}, we expect to find weakened decoder architectures for MIM to indirectly enhance the efficiency of the encoder when performing downstream tasks. 
The weaker decoder also helps to reduce the number of parameters of CMAE and optimize its efficiency. We designed four scenarios for the weakened decoder.
\begin{itemize}
  \vspace{-5pt}
\item[1)] Lowering parameters. We consider both cases of decoder layer reduction and decoder dimension reduction. We tested CMAE by reducing two layers per 	decoder. 
On the other hand we have designed a series of dimensions such as [512,256,128,64] for the decoder.
\vspace{-5pt}
\item[2)] MLP decoder. We replace all the transformer architectures in the MIM decoder	with MLP for testing. The decoder dimension is same as the corresponding MIM decoder dimension. The decoder depth is kept as the corresponding MIM 	decoder depth.
\vspace{-5pt}
\item[3)] Convolutional decoder. We designed a deep convolutional model with 	convolutional kernel of 3, step size of 1 and padding of 1 as the convolutional decoder of CMAE. 
The number of channels and depth of the 	convolutional decoder are aligned with the dimensionality and depth of the 	corresponding MIM decoder.
\vspace{-5pt}
\item[4)] Hybrid decoder. We try to degrade the decoder performance by mixing the 	transformer decoder in MIM with MLP and convolution respectively. We replace 	the decoder middle layer while keeping the original MIM decoder hyperparameters. 
Deep convolution contributes to the MIM decoder's ability 	to improve channel and local modeling. It also provides multi-scale information for the decoder in MIM. 
MLP provides a simpler architecture compared to transformer, which helps to improve decoder efficiency.
\end{itemize}
We still use the MSE loss as reconstruction loss $L_{con}$ to recover the mask token by reconstructing the pixel form.

\subsection{Objective function}
The loss function of CMAE consists of a contrast loss $L_{ctr}$, a location prediction loss $L_{loc}$, and a pixel reconstruction loss $L_{con}$. we can finally formulate our loss as
\begin{equation}
L = {\lambda _{ctr}}{L_{ctr}} + {\lambda _{{\rm{loc}}}}{L_{{\rm{loc}}}} + {\lambda _{con}}{L_{con}}
  \label{eq6}
  \end{equation}

$\lambda_{ctr}$,$\lambda_{loc}$ and $\lambda_{con}$ are the hyperparameters of contrast loss,location prediction loss and reconstruction loss,respectively. We default them all to 1.

\section{Experiments}
In this section we conduct experiments to verify the efficiency of CMAE. We first describe the data set and comparison methods in section 4.1. We then describe the experimental procedure in section 4.2. 
Then, we compare CMAE with other methods and show the experimental results in section 4.3.

\subsection{Dataset and Baseline Methodology}

We compared it with varies state-of-the-art self-supervised algorithms on tiny-imagenet. Tiny ImageNet contains 100000 images of 200 classes (500 for each class) downsized to 64×64 colored images. 
Each class has 500 training images, 50 validation images and 50 test images. We compared CMAE with MOCO-v1 \cite{he2020momentum}, MOCO-v2 \cite{chen2020improved}, and MAE \cite{he2021masked}.

\subsection{Experimental Details}
For MOCO-v1, we used ResNet18 as its base model, pre-trained with 500 epochs and the batch size set to 512. Then linear classification was performed using 100 epochs.

For MOCO-v2, we use ResNet50 for pre-training of 200 epochs in a batch size of 128. Then freeze the pre-trained model and perform linear classification with 100 epochs.

For MAE, we use ViT-Base as the base model in tiny-imagenet for 300 epochs of pre-training. MAE uses a mask ratio of 0.75 and a weight decay rate of 0.05. The default batch size is 64 and the base learning rate is 1e-3. 
The scheduled solution is cosine annealing. Next, the pre-trained model is frozen and fine-tuned for 100 epochs.

For CMAE, the details of our experiments are all consistent with MAE.

\subsection{Experimental Results}
\setlength\tabcolsep{1pt}

\begin{table}[htbp]
\tablestyle{5pt}{1.05}
\caption{Classification results of the self-supervised model on tiny-imagenet.}
\begin{tabular}{llccc}
    \toprule
    Method & Architecture & Epochs & Batch & Top-1 accuracy \\
    \midrule
    MOCO-v1 & \scriptsize ResNet-18 & 500   & 512   & 47.09\% \\
    MOCO-v2 & \scriptsize ResNet-50 & 200   & 128   & 45.82\% \\
    MAE   & \scriptsize ViT-B & 300   & 64    & 62.95\% \\
    CMAE  & \scriptsize ViT-B & 300   & 64    & \textbf{65.84}\% \\
    \bottomrule
    \end{tabular}%
  \label{tab1}%
\end{table}%

Our results on tiny-imagenet are shown in Tab.\ref{tab1}.

\section{Conclusion}
For better optimization of transformer training in self-supervised learning, we have discussed and investigated on the basis of MIM. We specifically studied three directions of improvement of the MIM paradigm. 
First we studied the improvement of the MIM input. We have introduced ContrastiveCrop into MIM in order to make MIM better for the purpose of recovering the main objects of images. 
And we have also explored how ContrastiveCrop can be better used for transformer. Second, we discussed the task of enhancing MIM. 
We introduced a contrastive task and a position prediction task for the MIM encoder to enhance the MIM's ability to perform downstream tasks. 
Third, we explore how the MIM decoder can be weakened so that indirect MIM can improve encoder performance. In terms of the results of tiny-imagenet, our study has achieved good results that provide other researchers with an in-depth exploration of MIM.

We will continue to update this paper, so stay tuned.

{\small
\bibliographystyle{ieee_fullname}
\bibliography{egbib}
}

\end{document}